\pdfoutput=1

\documentclass[11pt]{article}

\usepackage[]{emnlp2021}

\usepackage{times}
\usepackage{latexsym}

\usepackage[T1]{fontenc}

\usepackage[utf8]{inputenc}

\usepackage{booktabs}
\usepackage{multirow}
\usepackage{graphicx}

\usepackage{microtype}

\title{Investigating Robustness of Dialog Models to Popular Figurative Language Constructs}

\author{Harsh Jhamtani \thanks{\quad HJ and VG contributed equally for this paper. Order decided by coin flip.}\hspace{0.5em} $^1$ \quad Varun Gangal $^*$ $^1$ \quad  Eduard Hovy $^1$ \quad Taylor Berg-Kirkpatrick $^2$ \\
$^1$ School of Computer Science, Carnegie Mellon University\\
$^2$ Computer Science and Engineering. University of California San Diego \\
\tt{\{jharsh,vgangal,hovy\}@cs.cmu.edu, tberg@ucsd.eng.edu}
}

\date{}

\begin{document}
\maketitle
\begin{abstract}
Humans often employ figurative language use in communication, including during interactions with dialog systems. Thus, it is important for real-world dialog systems to be able to handle popular figurative language constructs like metaphor and simile. In this work, we analyze the performance of existing dialog models in situations where the input dialog context exhibits use of figurative language. We observe large gaps in handling of figurative language when evaluating the models on two open domain dialog datasets. When faced with dialog contexts consisting of figurative language, some models show very large drops in performance compared to contexts without figurative language. We encourage future research in dialog modeling to separately analyze and report results on figurative language in order to better test model capabilities relevant to real-world use. Finally, we propose lightweight solutions to help existing models become more robust to figurative language by simply using an external resource to translate figurative language to  literal (non-figurative) forms while preserving the meaning to the best extent possible.
\end{abstract}

\section{Introduction}

\begin{figure}[t]
    \centering
    \includegraphics[width=0.45\textwidth]{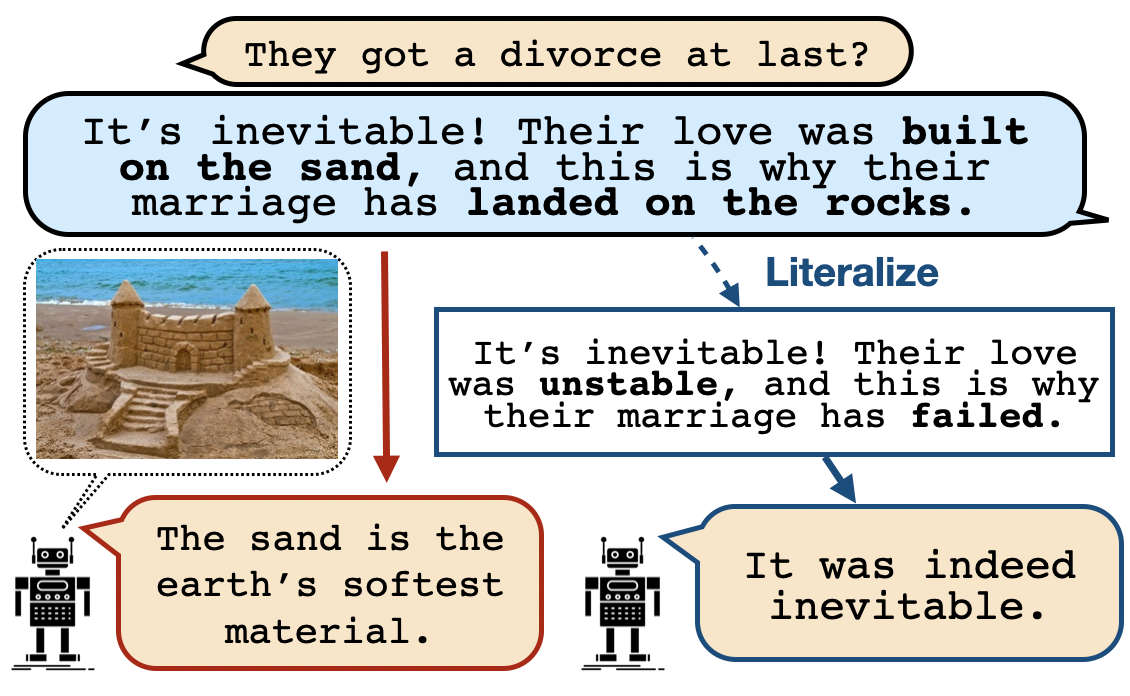} 
    \caption{\footnotesize
An example illustrating how model responses are affected by figurative constructs in dialog context. Here, the model conflates the metaphorical use of \textit{build on the sand} with its literal meaning, leading to an inappropriate, atopical response.}
    \label{fig:pullexample}
    \vspace{-3ex}
\end{figure}

Human frequently employ figurative language such as metaphors \cite{carbonell1982metaphor} and idioms \cite{jackendoff1995boundaries} for effective and/or stylistic communication. 
Thus, dialog models interacting with humans should be equipped to handle these forms of communication. 
However, understanding figurative language might be challenging for machines since figurative constructions often exhibit non-compositional semantics and may rely on shared cultural and common-sense knowledge \cite{carbonell1983metaphor}. For example, a powerful GPT2 model fine-tuned on DailyDialog dataset is unable to handle the metaphor `built on the sand' (Figure \ref{fig:pullexample}), and the response seems to rely on the unintended literal sense of `sand'. 

In this work, we investigate the performance of existing dialog models when faced with inputs containing figurative language use.
(1) First, we identify the subsets in existing datasets (such as DailyDialog \cite{li2017dailydialog}  and PersonaChat \cite{zhang2018personalizing}) which have figurative language use such as metaphors and similes. We observe that the performance of all the dialog models under consideration is lower on such subsets containing figurative language use compared to the dataset as a whole. 
(2) Second, we gather manually written literal/non-figurative equivalents of the dialog utterances in DailyDialog and PersonaChat which exhibit figurative language use. For example, literal equivalent of `on the sand' can be `unstable' (Figure \ref{fig:pullexample}). We observe that performance of dialog models improves when using literal equivalents in place of figurative language. We release the resulting datasets, and encourage that new dialog models be tested separately on such datasets to understand and measure their ability to handle figurative language. 
(3) Finally, we propose a simple defense against occurrences of figurative language in dialog context. More specifically, we use existing classifiers to detect presence of certain types of figurative language in dialog contexts, and use dictionary lookups to transform them to their literal counterparts before feeding them to the dialog models. The proposed technique is  lightweight, does not require any retraining of the models, and is effective --  though gaps still remain, leaving scope for interesting future explorations. \footnote{Data and code can be found at \url{https://github.com/vgtomahawk/Dialog-Fig-Speech-Robust}.}

\section{Figurative Language In Open Domain Dialog}

We experiment with \textbf{DailyDialog (DD)} dataset \cite{li2017dailydialog}, which is an open domain dialog corpus with 13.1K conversations on colloquial topics like Tourism, Health etc, of which 1K dialogs (6.74K utterances) form the test split.
To carry out the desired analysis, we need to first identify the utterances in the dataset which have figurative language use. To achieve high precision labeling, we rely on manual annotations instead of using external figurative language detectors/classifiers. The task was performed manually by two graduate students (native English speakers) studying in a university with English as the primary language of instruction. Additionally, we request the annotators to also  write down the literal equivalent versions of the utterances containing figurative language.
We release the resulting subset of DailyDialog dataset as DailyDialog-Figurative (\textbf{DD-Fig}), consisting of those dialog instances which contain figurative language, along with two manually written literal versions of each utterance.

Though figurative constructs are only mildly frequent at an utterance level (2.2\% in DD), their frequency of occurring at least once in a dialog is $\approx$6 times higher (13.1\%). This means that model sensitivity to figurative constructs is \emph{more significant} an issue than \emph{mere utterance-level frequency suggests}, since figurative constructs occurring anywhere in a dialog potentially affect all model responses for that dialog. 
Additionally, handling figurative constructs is still critical to robust long-tailed performance \cite{bamman2017natural}, which matters for worst-case behaviour and user satisfaction \cite{goel2010anatomy,ilievski2019identity}.
Lastly, most the prevalent figurative constructs we observe are metaphors (45.8\%), idioms (47.3\%), rhetorical questions (5.3\%), hyperbole (6.9\%), and personification (3.8\%).
Appendix \S B explores further properties of figurative prevalence.

\section{Experiments}

\begin{table}[t]
\centering
\small 
\begin{tabular}{@{}c@{\hskip 0.1in}c@{\hskip 0.1in}c@{\hskip 0.1in}c@{\hskip 0.1in}c@{\hskip 0.1in}c@{}}
 \textbf{Metric} & \textbf{Model} & \textbf{DD} & \textbf{DD-Fig} & \textbf{\% Drop} & \textbf{Rank-}  \\ 
  &  &  &  &  & \textbf{change}  \\ 
\toprule
Bleu-4 & GPT2 & 0.005 & 0.000 & 100.00 & 1\\
Bleu-4 & CVAE & 0.012 & 0.000 & 100.00 & -2 \\
Bleu-4 & DualEnc & 0.003 & 0.000 & 100.00 & 0 \\
Bleu-4 & HRED & 0.017 & 0.009 & 50.00 & -1 \\
Bleu-4 & Seq2Seq & 0.012 & 0.015 & -23.53 & 2 \\
\midrule
Meteor & GPT2 & 0.130 & 0.105 & 18.84 & -2 \\
Meteor & CVAE & 0.135 & 0.114 & 15.54 & -1 \\
Meteor & DualEnc & 0.111 & 0.105 & 5.31 & 1 \\
Meteor & HRED & 0.134 & 0.114 & 15.24 & -1 \\
Meteor & Seq2Seq & 0.121 & 0.115 & 5.61 & 3 \\
\midrule
Rouge-L & GPT2 & 0.166 & 0.146 & 11.90 & 0 \\
Rouge-L & CVAE & 0.319 & 0.283 & 11.47 & -1 \\
Rouge-L & DualEnc & 0.233 & 0.214 & 8.48 & 0 \\
Rouge-L & HRED & 0.329 & 0.297 & 9.91 & -1 \\
Rouge-L & Seq2Seq & 0.315 & 0.302 & 4.10 & 2 \\
\midrule
Human & GPT2 & 3.278 & 2.712 & 17.27\% & 0 \\
Human & CVAE & 2.302 & 1.771 & 23.06\% & -1 \\
Human & DualEnc & 1.699 & 1.397 & 17.79\% & 0 \\
Human & HRED & 2.353 & 2.115 & 10.10\% & 0 \\
Human & Seq2Seq & 2.146 & 1.870 & 12.85\% & 1 \\
\bottomrule
\end{tabular}
\caption{\small
We compare the model response quality for DD-Fig subset compared to the full DD. Across most of the models, we observe a considerable drop in response quality. 
Additionally, we note that relative ranks of various models as per automated metrics on Daily Dialog dataset show major changes when evaluating just on the figurative subset -- i.e. the best model overall might not be the best on the figurative subset, highlighting the need to separately report results on the data subset which has figurative language use.
}
\label{tab:dailydialogsubset}
\end{table}

\noindent \textbf{Do the dialog models perform worse on the data subset having figurative language use ?}
We compare model performances on DD vs DD-Fig. For DailyDialog data, we consider model outputs provided by \citet{gupta2019investigating} from the following methods: CVAE \cite{DBLP:conf/acl/ZhaoZE17}, HRED \cite{Serban2016BuildingED}, Seq2Seq \cite{DBLP:journals/corr/VinyalsL15}, Dual-encoder (DualEnc) \cite{DBLP:conf/sigdial/LowePSP15}, and GPT2-medium \cite{radford2019language} fine-tuned on DD.
To report automated metrics, we use the multi-reference annotation set collected by \citet{gupta2019investigating}. 
However, automated metrics may not be well correlated with output response quality \cite{DBLP:journals/corr/abs-2008-12009,DBLP:conf/acl/GangalJHB21}. Therefore, we also carry out human evaluations, wherein human annotators (on Amazon Mechanical Turk) are asked to judge the appropriateness of a dialog response on a 1-5 Likert scale.
We observe that most of the models perform much worse on DD-Fig (Table \ref{tab:dailydialogsubset}), with the drop being close to 99\% in some cases.  \vspace{2mm}

\noindent \textbf{Relative Performance of Models:}
We additionally investigate if the relative ranks of the models change when evaluating only on the subset with figurative language use compared to the entire test split. We note that there are some substantial changes in the relative ranks of the models (Table \ref{tab:dailydialogsubset}). For instance, as per Meteor and human-ratings, Seq2Seq performs better than CVAE on the figurative subset, while doing worse on the complete test dataset. Such changes in relative ranks further highlight the need to separately report results for the proposed data subsets. 
Interestingly, Seq2Seq improves its relative rankings in general on the DD-Fig subset. We hypothesize this is because it often generates very generic responses with 
little regard to the 
the context.
\vspace{2mm} 

\noindent \textbf{Does replacing figurative language with semantically equivalent literal translations lead to better performance?}
Above analysis only reports correlation in performance with contexts containing figurative language. However, there could be certain confounding factors involved. Thus, to make a more direct comparison, we compare model generated response when using figurative contexts versus their literal counterparts. To perform this experiment, we utilize the human written literal versions in DD-Fig, and experiment with the GPT2 model (which is the best performing model as per human rating on the overall dataset). 
We report results under two setups: (1) when figurative language is present in the last utterance of the dialog history, and (2) when figurative language is present anywhere in the dialog history. Human ratings are collected using the same procedure as described for Table \ref{tab:dailydialogsubset}.

\begin{table}[t]
\centering
\small 
\begin{tabular}{@{}lcccc@{}}
\bf Data/Model: & \bf DD-Fig/GPT2 & \bf PC-Fig/GPT \\
\toprule
\multicolumn{3}{l}{Figurative in last utterance} \\
Bleu-3/4 & +98\% / +477\% & NA  \\
Meteor/Rouge-L & +5.3\% / -3.6\% & NA  \\
Human-rating & +12.1\% & +17.2\% \\
\midrule
\multicolumn{3}{l}{Figurative anywhere in dialog history} \\
Bleu-3/4 & +13\% / +79\% & NA\\
Meteor/Rouge-L & +3.9\% / -3.0\% & NA  \\
Human-rating & +13.7\% & +19.5\% \\
\bottomrule
\end{tabular}
\caption{\small
We compute the \% change in performance when figurative language replaced with manually written literal counterparts in DD-Fig and PC-Fig data subsets.
We observe that replacing figurative text in dialog context with literal version leads to improved dialog response quality.
}
\label{tab:dailydialogparallel}
\vspace{-4mm}
\end{table}

Table \ref{tab:dailydialogparallel} shows the main results. 
For some metrics such as Bleu-4, models perform more than 5 times better when fed with literal translations instead of figurative language. 
Between the two setups under consideration, we observe slightly higher impact (as per human evaluation ratings) when one or more figurative language constructs are in use anywhere in the dialog history. 
\vspace{2mm}

\noindent \textbf{Experiments with Personachat (PC):}
PersonaChat \cite{zhang2018personalizing} is a persona grounded dialog corpus, with 1K dialogs (7.8K utterances) forming the test split. 
We use pre-trained GPT model and fine-tune it on the train split of the PC dataset.
We follow the same tagging and annotation procedure for the test split of PC as we did on DD, and refer to the resulting dataset as PersonaChat-Figurative dataset (\textbf{PC-Fig}). Results in Table \ref{tab:dailydialogparallel} demonstrate reduction in performance of the dialog model in human evaluations (Automated overlap metrics in the case of PC are considered unreliable since PC contains only one reference per dialog context). 

We notice that compared to DailyDialog, PersonaChat utterances tend to be shorter, more informal, and highly spoken-language like utterances, with fast topic transitions. On replacing figurative language with its literal counterpart into such contexts, the replaced literal text, which is typically English of a more formal and written variety, ends up being much more out of sync with the context in the lexico-syntactic/stylistic sense than it is for DailyDialog. This has a slight downward effect on metrics, offsetting some of the gains from replacing away the figurative language.

\section{Mitigation}

We propose a lightweight mitigation approach wherein we use existing resources to detect and then construct literal translations for two popular figurative constructs: metaphors and idioms. 
Thus, the proposed mitigation approach does not require any retraining of dialog models.
\vspace{1mm}

\noindent{\textbf{Metaphor Detection Through Classifier}:} 
We train a metaphor detection classifier based on the VUA corpus \cite{steen2010metaphor,gao2018neural}\footnote{https://github.com/gao-g/metaphor-in-context}. 
 To better generalize to external data via recent contextual models, we skip using  model by \citet{gao2018neural}, and instead learn a classifier $C^{met}_{bert}$ by finetuning the \textit{bert-base-uncased} \cite{devlin2018bert} checkpoint from \citet{wolf2019huggingface}. On VUA, $C^{met}_{bert}$ gets a test F1 of $0.724$, which is close to \citet{gao2018neural}'s value of $0.726$. 
Next, we run each test utterance in dilaog dataset through $C^{met}_{bert}$ to get its probability $p^{met}$ of being metaphorical. To retain only more reliable predictions, especially considering domain shift w.r.t VUA, we only choose utterances with $p^{met}>$0.9. The set of metaphorical utterances thus identified is $D^{met}_{auto}$. \vspace{2mm}

\noindent{\textbf{Idiom Detection Through Lexicon:}} 
Idioms are frequently used expressions, which have a fixed, typically non-compositional meaning understood by most native speakers from cultural precedent. We curate a lexicon of $2048$ commonly used idioms (e.g. \textit{filling his shoes}) from an online source\footnote{https://www.englishclub.com/ref/Idioms/} -- see Appendix A for more details). All utterances with at least one lexicon entry as a substring are identified to create the set of automatically detected idiomatic utterances, $D^{idiom}_{auto}$. 
We unify the sets detected above to form $D^{fig}_{auto}= D^{met}_{auto} \cup D^{idiom}_{auto}$. $|D^{fig}_{auto}|$ constitutes 1520 of 6740 utterances for \textbf{DD} (22.5\%) and 911 of 7801 utterances for \textbf{PC} (11.7\%) respectively. \vspace{2mm}

\noindent{\textbf{Dictionary Replacement:}} 
Wiktionary \cite{zesch2008extracting} - the collaboratively created online dictionary, provides a curation of entries corresponding for phrases with ``idiomatic"\footnote{Overloaded use of the term to refer to several figurative phenomena at once, and not just idioms proper.} usages. These entries encompass conventionalized metaphors \footnote{Commonly used metaphors with a fixed, nearly universally accepted meaning.}, idioms, euphemisms, commonly used similes etc. 
Each entry lists the surface form of the figurative construct paired with a gloss. Glosses are for the most part literal interpretations of the figurative construct.
However, they often bear other details like dialect(``\textit{US}''), etymology(``\textit{archaic}'') etc, which we remove through simple regex-based rules. This allows direct use of the now-cleaned gloss as a literal interpretation in-context.  
Furthermore, we expand entries whose surface forms contain uninflected verb forms or unrealized pronouns indicated by \textit{someone}, \textit{one's} etc, spawning one new entry per pronoun-inflection combination \footnote{We use \emph{pyinflect} python library} 
This yields us a dictionary with $17,743$ tuples of the form $\{fig_{i},Lit(fig_{i})\}$ \footnote{See Appendix \S C for further analysis of the dictionary}. 
Finally, for each detected utterance $u \in |D^{fig}_{auto}|$, each matched occurrence of $fig_{i}$ is replaced by $Lit(fig_{i})$, $\forall 1 \leq i \leq n$. \vspace{2mm}

\noindent \textbf{Results:}
From Table \ref{tab:mitigationparallel}, we see that mitigation-based literalization leads to higher quality model responses as per most automatic metrics as well as human evaluation.
Though the proposed approach offers only small improvements, it is lightweight in terms of time and memory complexity, and provides reasonably fluent and appropriate interpretations for the figurative constructs covered, since these are sourced from the long-term, collaborative editing underlying Wiktionary. Table \ref{tab:mitigationexamples} shows examples where mitigation-based literalization of the figurative context improves model response quality. 

Additionally, we observe that Rouge fails
to correlate with Meteor in Table \ref{tab:dailydialogparallel}, but correlates in Table \ref{tab:mitigationparallel}. One possible reason for such behavior is that Wiktionary uses dictionary-like, conservative literalizations, adding new words only as necessary. On the other hand, human annotators literalize more freely without regard for word choice fidelity. Meteor is more robust to variation in word choice, being enabled to capture synonymy and other forms of limited surface form variation. Rouge, being more sensitive however, is immediately dampened on account of this.

\begin{table}[t]
\centering
\small 
\begin{tabular}{@{}lcccc@{}}
\bf Data/Model: & \bf DD-Fig/GPT2 & \bf PC-Fig/GPT \\
\toprule
\multicolumn{3}{l}{Figurative in last utterance} \\
Bleu-3/4 & +46.3\%/+54.1\% & NA  \\
Meteor/Rouge-L & +2.0\%/+5.7\% & NA \\
Human-rating & +13.7\% & +0.4\% \\
\midrule
\multicolumn{3}{l}{Figurative anywhere in dialog history} \\
Bleu-3/4 & +8.1\%/+0.6\% & NA \\
Meteor/Rouge-L & +1.4\%/+1.4\% & NA \\
Human-rating & +7.7\% & +0.3\% \\
\bottomrule
\end{tabular}
\caption{\small
Mitigation: Contrasting performance of GPT-based models on the (automatically detected) subset $D^{fig}_{auto}$, containing figurative language, against the same subset but with figurative language replaced with \emph{Wiktionary-mitigated} literal counterparts.
}
\label{tab:mitigationparallel}
\vspace{-4mm}
\end{table}

The proposed approach is based on simple rule based procedures relying on existing resources, and thus there is scope of multiple future extensions.
The detection portion of our approach uses an external classifier  and a fixed lexicon to detect metaphors \& idioms respectively, leading to $D^{met}_{auto}$. Considering utterances in DD-Fig as gold, we find the recall of this approach, given by 
$\|\frac{D^{met}_{auto} \cap \textrm{\footnotesize DD{-}Fig}\|}{\|\textrm{\footnotesize DD-Fig}\|} = 0.31$ to be on the lower end. This maybe due to the detection 1) missing out on entirely creative \textit{novel} metaphors, rare idioms as well as other unhandled figures of speech like rhetorical questions. 2) being unable to pick out figures of speech from domain shifted contexts. Extending detection to a wider range of figurative constructs and improved ways to address domain shift are thus important avenues for future work.

Furthermore, the proposed mitigation approach is deficient in its ability to literalize all figurative text, since it can't handle: (1) Novel metaphors, which, being creative are unseen/infrequent in extant corpora, as opposed to \emph{conventionalized} metaphors. (2) Complex figurative constructs which lack particular surface forms and are context sensitive or highly creative - e.g sarcasm \& hyperbole.
Future work could look into improved methods for figurative language detection, perhaps including use of resources such as commonsense knowledge resources \cite{DBLP:conf/acl/BosselutRSMCC19} which have been separately shown to be useful in dialog domain \cite{DBLP:conf/emnlp/MajumderJBM20} and in generating figurative language \cite{chakrabarty2020generating}.

\begin{table}[t]
\centering
\footnotesize
\begin{tabular}{@{}l@{}}
\toprule
\textsc{Figurative Context (PC)} \\ 
maybe we can \textit{get together} sometime \\ if you are not scare of a 30 year old cougar ! \\  \midrule
\textsc{Model Response:} \\ 
i'm not scared of any \textit{cats}. i've two dogs \\  \midrule
\textsc{Literalized Context} \\ 
maybe we can start dating sometime \\ if you are not scare of a 30 year old cougar ! \\  \midrule
\textsc{Model Response:}  \\
that's a very interesting idea. i've never met one. \\ \midrule \midrule
\textsc{Figurative Context (DD)} \\ 
I think it's time for me to meet my admirer and \\make him \textit{face the music} . \\  \midrule
\textsc{Model Response} \\ 
let's \textit{rock and roll}. \\  \midrule
\textsc{Literalized Context} \\ 
I think it's time for me to meet my admirer and \\ make him bear the consequences of his actions . \\  \midrule
\textsc{Model Response:}  \\
let's meet. \\ 
\bottomrule
\end{tabular}
\caption{
Examples from PersonaChat(PC) and DailyDialog (DD) illustrating how figurative contexts on literalization produce better (GPT-2) model responses. See Appendix F for additional examples.
}
\label{tab:mitigationexamples}
\end{table}
\section{Related Work}

Past work has explored fine-grained analysis and understanding of the performance of dialog models  \cite{DBLP:journals/corr/abs-2006-12442}. 
\newcite{DBLP:journals/corr/abs-2006-08331} analyze open domain dialog systems for skills such as inferring contradictions and determining the topic of conversation inter alia. 
\newcite{DBLP:conf/acl/SankarSPCB19} analyze the change in perplexity when applying certain perturbations in dialog history. 
Past work has analyzed dialog models from the point of view of safety from toxic language \cite{DBLP:journals/corr/abs-2010-07079,DBLP:conf/emnlp/DinanHCW19}, and gender biases \cite{dinan2020queens}. \citet{gao2020dialogue} analyze how well dialog models respond to utterances from infrequent sentence function types (e.g \textit{Negative Declarative} utterances like \textit{I feel bad today.}). 
\citet{DBLP:conf/emnlp/LouisRR20} propose to identify the categorical mapping of an indirect response with respect a polar question in a task oriented dialog setup. 

Challenges in handling metaphors and idioms has been explored in prior work on machine translation \cite{DBLP:conf/starsem/MohammadST16,DBLP:conf/acl/Kordoni18,DBLP:conf/acl/LinGM18}.
\citet{DBLP:conf/acl/LinGM18} propose a method to identify metaphors in English text, and paraphrase them into their literal counterparts before translating to Chinese. Our work on analyzing dialog models against figurative language contexts is along similar direction, though the task setup and scope of figurative language involved are different.  
Figurative language generation has received reasonable attention such as simile generation \cite{chakrabarty2020generating} and idiom generation \cite{zhou2021solving}. Compared to them, our focus is on analyzing capability of popular contemporary dialog models when faced with figurative language. 
\section{Conclusions}
In this work, we demonstrate how existing dialog models fall short in handling figurative language use, and propose a light-weight mitigation technique to ameliorate this lacuna. 
We encourage future research in dialog models to separately analyze and report model performance 

The mitigation techniques used by us are pretty lightweight, but are not able to capture many occurrences of figurative language. Future work could look into improved techniques for figurative language detection. Our work is limited to a couple of open domain dialog datasets in English language. Similar analyses could be conducted on goal oriented dialog setups and datasets in other languages.

\section*{Acknowledgements}
We thank anonymous EMNLP reviewers for insightful comments and feedback.

\section*{Ethics Statement}
Our human preference/appropriateness ratings are collected over source content either directly sourced from, or based on typical, off-the-shelf models trained  on already existing, publicly available and widely used dialog datasets - namely, DailyDialog  \cite{li2017dailydialog} and Personachat \cite{zhang2018personalizing} as well as the multiple references dataset from \cite{gupta2019investigating}. 
We do collect human evaluation ratings using crowd-sourcing. However, we neither solicit, record or request any kind of personal or identity information from the annotators.
Our work  primarily performs experiments on dialog in English \cite{bender2018data}.
Dialog models are known to suffer from biases learnable from dialog training data, such as gender bias \cite{dinan2020queens}. However, our work and contribution does not present or release any new models or model checkpoints, and is primarily concerned with more careful evaluation of a particular phenomena (i.e figurative language), and discussion on lightweight mitigation strategy related to the same.

\bibliographystyle{acl_natbib}
\bibliography{acl2021}

\clearpage
\appendix

\section{Idiom Detection}
Lexicon construction is done through a two step process. First, we curate the lexicon from the mentioned source. Since many entries in this lexicon are templates like \textit{behind someone's back} which could have multiple realizations (e.g \textit{behind her back}), we use a rule-based procedure to expand such instances to all possible realizations by enumerating over all POS combinations (nominative pronoun, nominative pronoun-verb, verb-accusative pronoun etc as applicable). For this, we use the \textit{pyinflect} library.

\section{Additional Discussion on Properties of Figurative Utterances} \label{subsec:propfigcontinued}
\noindent \textbf{Dialog Acts:}
Since DailyDialog has dialog acts annotated amongst 4 types ( \textit{inform}, \textit{question}, \textit{directive}, \textit{commissive}), we analyze how figurative language distributes over these types. Amongst figurative utterances, \textit{inform} is more dominant than overall (58.7\% vs 43.2\%), while \textit{question} (32.2\% vs 22.0\%) and \textit{directive} (10.7\% vs 17.3\%) are underrepresented.

\section{Wiktionary Mitigation - Resource Description} \label{sec:wiktionaryAnalysis}
The final postprocessed version of the resource contains 17743 entries - these are distributed as 88.53\% `idioms', 8.08\% euphemisms and 3.39\% similes. The `idioms' here consist of both idioms proper and conventionalized metaphors - it is not easy to provide an exact breakup since Wiktionary does not distinguish between the two.

\subsection{Examples}
\label{subsec:wiktionaryExamples}
In Table \ref{tab:wiktionaryExamples}, we enlist a few examples from the Wiktionary lexicon for each figurative construct type. The complete resource can be found in our submission materials at \textit{Data/MitigationDictionary} 

\begin{table}[t]
\centering
\small
\begin{tabular}{@{}lll@{}}
Type & Phrase & Literalization \\
\toprule
Idiom & a bridge too far &  a step or action that is \\ 
& & too ambitious  \\ 
\midrule
Idiom & a life of its &  an independent existence \\ 
& own &  with some characteristics  \\ 
& & of life \\
\midrule
Idiom & a wild goose &  most things are inherited \\ 
& never laid a & and predetermined  \\ 
&  tame egg & \\
\midrule
Euphemism & aurally &  deaf or hard of \\ 
& challenged & hearing  \\ 
\midrule
Euphemism & between jobs &  unemployed \\ 
\midrule
Euphemism & bite the dust &  die \\ 
\midrule
Simile & as modern as &  thoroughly \\
& next week & modern \\
\midrule
Simile & avoid like &  evade or \\
& the plague & shun \\
\midrule
Metaphor & bear the &  endure the worst \\
& brunt & part of something \\
\midrule
Metaphor & beat a &  persist or continue \\
& dead horse & far beyond any \\
&  & specific purpose \\
\midrule
Metaphor & cede the &  withdraw from any \\
& field & confrontational or \\
&  & potentially confrontational \\
&  & situation \\
\bottomrule
\end{tabular}
\caption{Examples from the Wiktionary lexicon resource}
\label{tab:wiktionaryExamples}
\vspace{-3ex}
\end{table}

\section{Additional Implementation Details}

\textbf{Compute Infrastructure:} We used Nvidia RTC 2080 Ti GPU cards to run GPU based models. \\

\noindent \textbf{Run time:} 
On an average, it took about 5 hours to train GPT2 based models on DD dataset.
Typical inference time ranges from 10-20 minutes. \\

\noindent \textbf{Number of parameters:} 
GPT2 has approx 120M parameters. \\

\noindent \textbf{Hyper-parameter search:} We varied random seed and the learning rate, when training GPT based models. We use validation loss to pick the best configuration. Best configuration uses initial random seed of 123.

\section{Additional Data Collection Details}

\begin{figure*}
    \centering
    \includegraphics[width=0.95\textwidth]{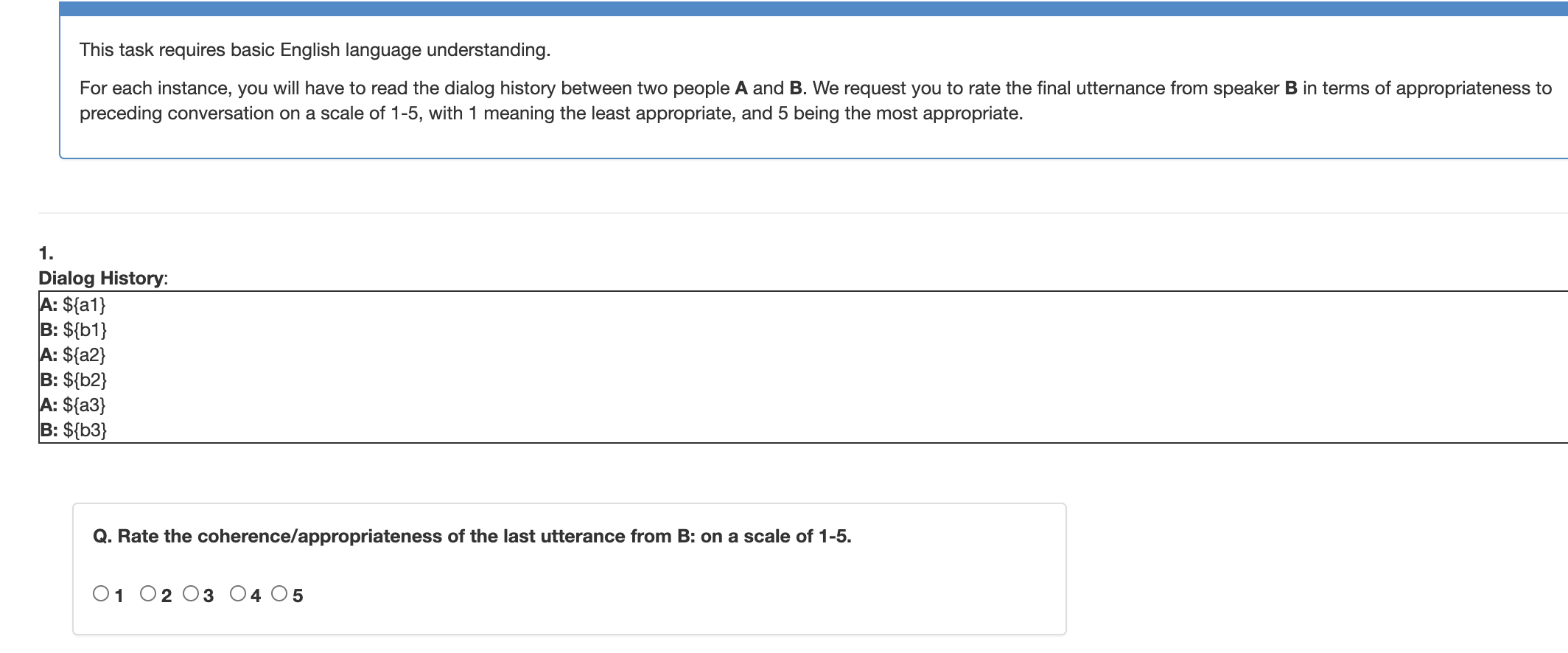}
    \caption{Annotation Framework to collect human judgement ratings on appropriateness of generated response with respect to dialog context.}
    \label{fig:amt}
\end{figure*}

\textbf{Annotation Framework:} 
A screenshot of the rating annotation collection task is shared in Figure \ref{fig:amt}.

\noindent \textbf{Quality Control}
We restrict to annotators with $>90\%$ HIT acceptance rate. We also perform spot checks and skip the ratings from annotators who seemed not to be adhering to the provided instructions. 

\section{Qualitative Examples}
In this section, we enlist qualitative examples from the various stages of analysis performed in our work. Table \ref{tab:qualExamples} shows how various models respond to example figurative contexts. Table \ref{tab:pullexamples} shows examples where figurative contexts are misinterpreted by models, with the response relying on the unintended literal sense of the construct. Table 
\ref{tab:mitigationexamples}
shows examples where literalization of the figurative context improves  quality of the model response. 

\begin{table}[t]
\centering
\footnotesize
\begin{tabular}{@{}l@{}}
\toprule
\textsc{Dialog Context:} \\ 
A: sure do . i \textit{get a kick out of} looking punk . \\  
B: this one looks really expensive though . \\  \midrule
\textsc{Human Written Response:} \\ 
A: it 's actually quite reasonably priced . \\  \midrule
\textsc{Model Response (GPT2):}  \\
A: i'm not one for a kick out. \\ \midrule \midrule
\textsc{Dialog Context:} \\ 
A: not good . my cousin is \textit{driving me up the wall} . \\  
B: how so ? \\  \midrule
\textsc{Human Written Response:} \\ 
A: he is always being so loud and annoying . \\  \midrule
\textsc{Model Response (GPT2):}  \\
A: you know i'm a good \textit{driver}. \\ \midrule \midrule
\textsc{Dialog Context:} \\ 
B: every time when heroes came back from the \\ battlefield  we \textit{took our hats off} to them . \\  \midrule
\textsc{Human Written Response:} \\ 
A: i think some people still do that today . \\  \midrule
\textsc{Model Response (GPT2):}  \\
A: they are just as beautiful as the \textit{red haired cap}. \\
\midrule \midrule
\textsc{Dialog Context:} \\ 
A: that dress is very pretty. why don't you like it ?  \\ 
B: \emph{It's too loud.} \\ \midrule
\textsc{Human Written Response:} \\ 
A: we've been looking around for many hours. \\  \midrule
\textsc{Model Response (CVAE):}  \\
A: i think it 's too noisy for you. \\
\bottomrule
\end{tabular}
\caption{
Examples from DailyDialog illustrating how various model responses are affected by figurative constructs. 
}
\label{tab:pullexamples}
\end{table}

\begin{table*}[ht!]
\centering
\scriptsize
\addtolength{\tabcolsep}{-4pt}
\begin{tabular}{|l|l|l| }
\hline 
No & Type  & Text \\ \hline 
\multirow{3}{*}{105-0} &   \textsc{Context} & {\color{red} A: How come it is slow as a snail today ?}  \\
& \textsc{GPT2} & {\color{teal} B: i think so. \textbf{it is a virus.} }  \\
&  \textsc{CVAE} &    {\color{brown} B: it 's a skill . i have to work out the nerve . } \\
& \textsc{DualEnc} & {\color{violet} B: we are basically going to check your heart , lungs , blood sugar levels , and eyes , ears , and nose .  } \\  
& \textsc{HRED} & {\color{blue} B: it 's a good idea .} \\
& \textsc{Seq2Seq} & {\color{blue} B: i 'm afraid you have to . } \\
\hline
\multirow{3}{*}{1-1} &   \textsc{Context} & {\color{red} A: What would the roses cost me ? B: They're only \$ 20 a dozen. A: I like the sound of that }   \\
& \textsc{GPT2} & {\color{teal} B: i 'll take one.}  \\
&  \textsc{CVAE} & {\color{brown} B: you can always get rid of the foods .} \\
& \textsc{DualEnc} & {\color{violet} B: you 'll only have to pay the initial fee . } \\ 
& \textsc{HRED} & {\color{blue} B: that 's a good idea . } \\
& \textsc{Seq2Seq} & {\color{orange} B: i 'll take it .} \\
\hline
\end{tabular}
\caption{Response utterances from five different models to some figurative contexts}
\label{tab:qualExamples}
\vspace{-3ex}
\end{table*}

\end{document}